\documentclass[conference]{IEEEtran}

\newif\ifanonymous        % 创建 \ifanonymous … \else … \fi 结构
\anonymoustrue            % ← 盲审版     （把它注释掉改成 \anonymousfalse 即为公开版）
\usepackage[switch]{lineno}
\IEEEoverridecommandlockouts
\usepackage{cite}
\usepackage{amsmath,amssymb,amsfonts}
\usepackage{algorithmic}
\usepackage{graphicx}
\usepackage{textcomp}
\usepackage{xcolor}
\usepackage{multirow}   % 多行单元格
\usepackage{booktabs}   % 若想用 \toprule 等请保留
\usepackage{array}      % 让列宽度更灵活（可选）
\usepackage{colortbl}
\usepackage{xcolor}
\def\BibTeX{{\rm B\kern-.05em{\sc i\kern-.025em b}\kern-.08em
    T\kern-.1667em\lower.7ex\hbox{E}\kern-.125emX}}
\begin{document}

\title{CafeMed: Causal Attention Fusion Enhanced Medication Recommendation\\
}

% ——— 作者信息（放在 \maketitle 之前） ———
% \ifanonymous
%   % ===========================
%   % 盲审版本：仅显示占位信息
%   % ===========================
%   \author{
%     \IEEEauthorblockN{Anonymous Authors}
%   }
% \else
  % ===========================
  % 公开版本：正式署名
  % ===========================
  % \author{
  %   \IEEEauthorblockN{1\textsuperscript{st} Kelin Ren}
  %   \IEEEauthorblockA{\textit{Hanyang University, Korea}\\
  %                     Ansan, Korea}
  %   \and
  %   \IEEEauthorblockN{2\textsuperscript{nd} Given Name Surname}
  %   \IEEEauthorblockA{\textit{Hanyang University, Affiliation}\\
  %                     Ansan, Korea}
  %   \and
  %   \IEEEauthorblockN{3\textsuperscript{rd} Given Name Surname}
  %   \IEEEauthorblockA{\textit{Hanyang University, Affiliation}\\
  %                     Ansan, Korea}
  % }
% \fi

\author{
  \IEEEauthorblockN{
    Kelin Ren$^{1}$,
    Chan-Yang Ju$^{2}$,
    Dong-Ho Lee$^{2}$$^*$\thanks{*Corresponding author.}
  }
  \IEEEauthorblockA{
    $^{1}$Department of Computer Science and Engineering, Hanyang University, Ansan, Korea \\
    $^{2}$Department of Applied Artificial Intelligence, Hanyang University, Ansan, Korea \\
    \{renkelin, karunogi, dhlee72\}@hanyang.ac.kr
  }
}

\maketitle

\begin{abstract}
Medication recommendation systems play a crucial role in assisting clinicians with personalized treatment decisions. While existing approaches have made significant progress in learning medication representations, they suffer from two fundamental limitations: (i) treating medical entities as independent features without modeling their synergistic effects on medication selection; (ii) employing static causal relationships that fail to adapt to patient-specific contexts and health states. To address these challenges, we propose CafeMed, a framework that integrates dynamic causal reasoning with cross-modal attention for safe and accurate medication recommendation. CafeMed introduces two key components: the Causal Weight Generator (CWG) that transforms static causal effects into dynamic modulation weights based on individual patient states, and the Channel Harmonized Attention Refinement Module (CHARM) that captures complex interdependencies between diagnoses and procedures. This design enables CafeMed to model how different medical conditions jointly influence treatment decisions while maintaining medication safety constraints. Extensive experiments on MIMIC-III and MIMIC-IV datasets demonstrate that CafeMed significantly outperforms state-of-the-art baselines, achieving superior accuracy in medication prediction while maintaining the lower drug--drug interaction rates. Our results indicate that incorporating dynamic causal relationships and cross-modal synergies leads to more clinically-aligned and personalized medication recommendations. Our code is released publicly at https://github.com/rkl71/CafeMed.
\end{abstract}


\section{Introduction}
With the growing imbalance between medical resource supply and demand, artificial intelligence technologies offer scalable solutions to automate clinical analysis and enhance decision-making efficiency \cite{yang2021safedrug, chen2023context, li2024stratmed}. Personalized medication recommendation has become a key application of EHR-based data mining \cite{cowie2017electronic, evans2016electronic}, where recent methods employ statistical modeling \cite{ma2018general, symeonidis2021recommending, wang2021learning} and deep learning \cite{shang2018knowledge, wang2019order, wang2022ffbdnet} to capture longitudinal patient histories \cite{lee2020clinical, shang2019gamenet}. However, substantial patient heterogeneity remains a major challenge: even individuals with 80\%–90\% similarity in clinical status may share less than 50\% overlap in actual medication usage \cite{li2024causalmed}, underscoring the necessity of highly personalized and context-aware medication recommendation.

Despite continuous progress, existing methods still face two primary limitations. First, most approaches rely on correlation-based associations among diseases, procedures, and medications, which may introduce spurious relationships and compromise clinical safety. Second, as illustrated in Fig. 1, many models encode diagnoses and procedures independently and fuse them only at the final prediction stage. This late fusion design overlooks synergistic effects and interactions across medical modalities, limiting accurate characterization of patient health states and reducing personalization quality.

\begin{figure}[!t]
  \centering
  \includegraphics[width=\linewidth]{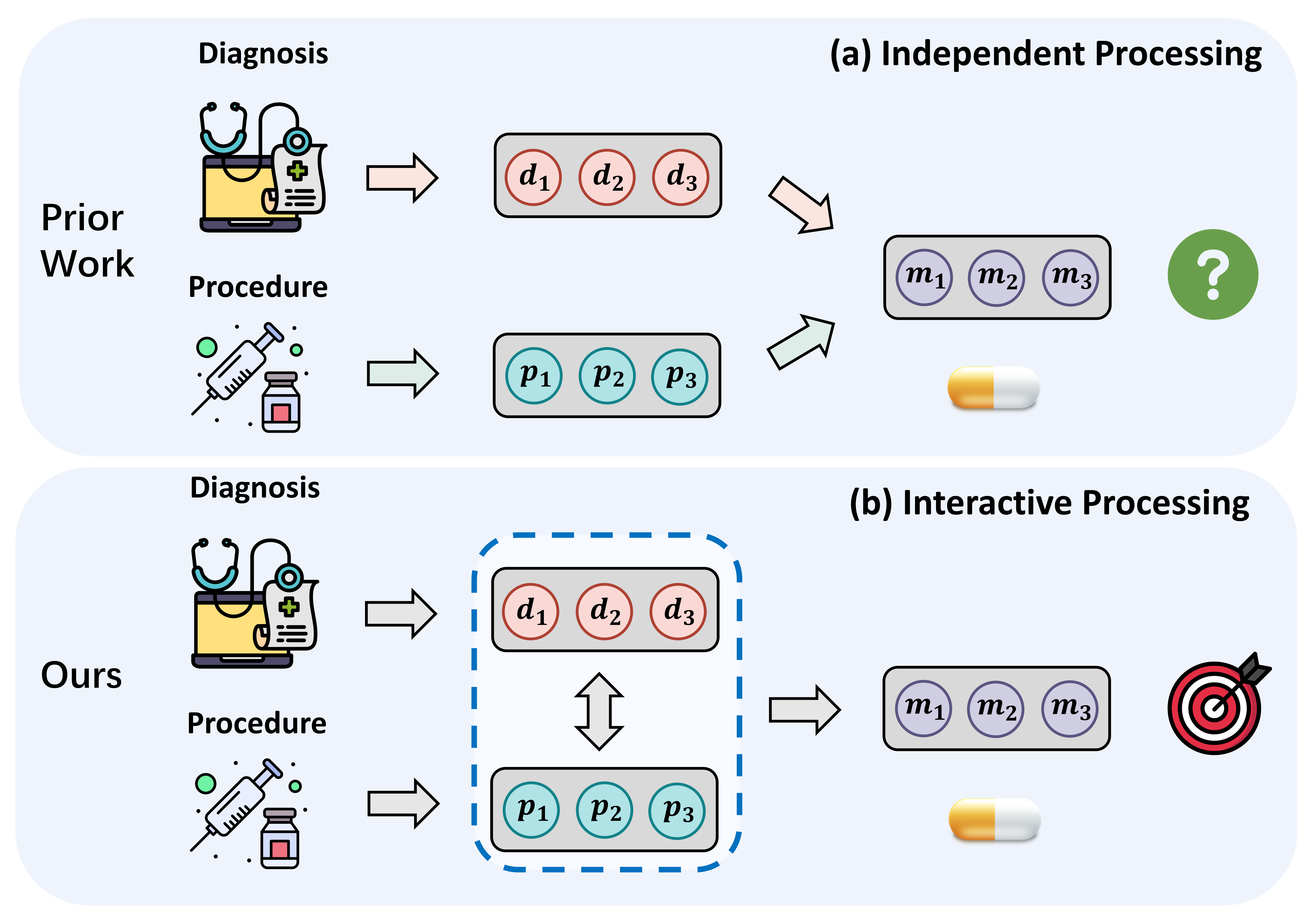}
  \caption{Comparison of causal modeling approaches in medication recommendation.  Prior methods (a) use independent processing pipelines where diagnosis and procedure information are encoded separately and merged only at the final prediction stage, potentially ignoring critical cross-modal interactions. Our proposed interactive processing framework (b) introduces early-stage feature interaction between diagnosis and procedure representations, enabling the model to capture synergistic effects and medical contraindications, resulting in more accurate medication recommendations.}
\end{figure}

To address these limitations, we propose CafeMed, a \textbf{C}ausal \textbf{A}ttention \textbf{F}usion \textbf{E}nhanced \textbf{Med}ication recommendation framework that integrates dynamic causal reasoning with cross-modal attention. CafeMed employs a \textbf{C}ausal \textbf{W}eight \textbf{G}enerator (CWG) to adaptively modulate causal strengths according to patient-specific contexts, and a \textbf{C}hannel \textbf{H}armonized \textbf{A}ttention \textbf{R}efinement \textbf{M}odule (CHARM) to capture fine-grained interactions between diagnoses, procedures, and medications. These components jointly enable safe, accurate, and personalized medication recommendation. Our main contributions are summarized as follows:

% To address the above problems, we propose a \textbf{C}ausal \textbf{A}ttention \textbf{F}usion \textbf{E}nhanced \textbf{Med}ication Recommendation named CafeMed. First, we design the \textbf{C}ausal \textbf{W}eight \textbf{G}enerator (CWG) mechanism, which dynamically modulates the causal interaction weights between diseases and procedures based on patient-specific health states, thereby precisely amplifying the medical factors most relevant to current therapeutic decisions. Second, we propose the \textbf{C}hannel \textbf{H}armonized \textbf{A}ttention \textbf{R}efinement \textbf{M}odule (CHARM), which employs multi-channel attention mechanisms to finely capture complex synergistic relationships among diseases, procedures, and medications, optimizing their interactive representations for enhanced recommendation accuracy. This dual strategy of causal gating and attention fusion enables CafeMed to more deeply understand and accurately portray the real health status of patients, thus significantly improving the personalization and safety of recommendation results.

% 在这里插入大图
\begin{figure*}[t]
  \centering
  \includegraphics[width=\textwidth]{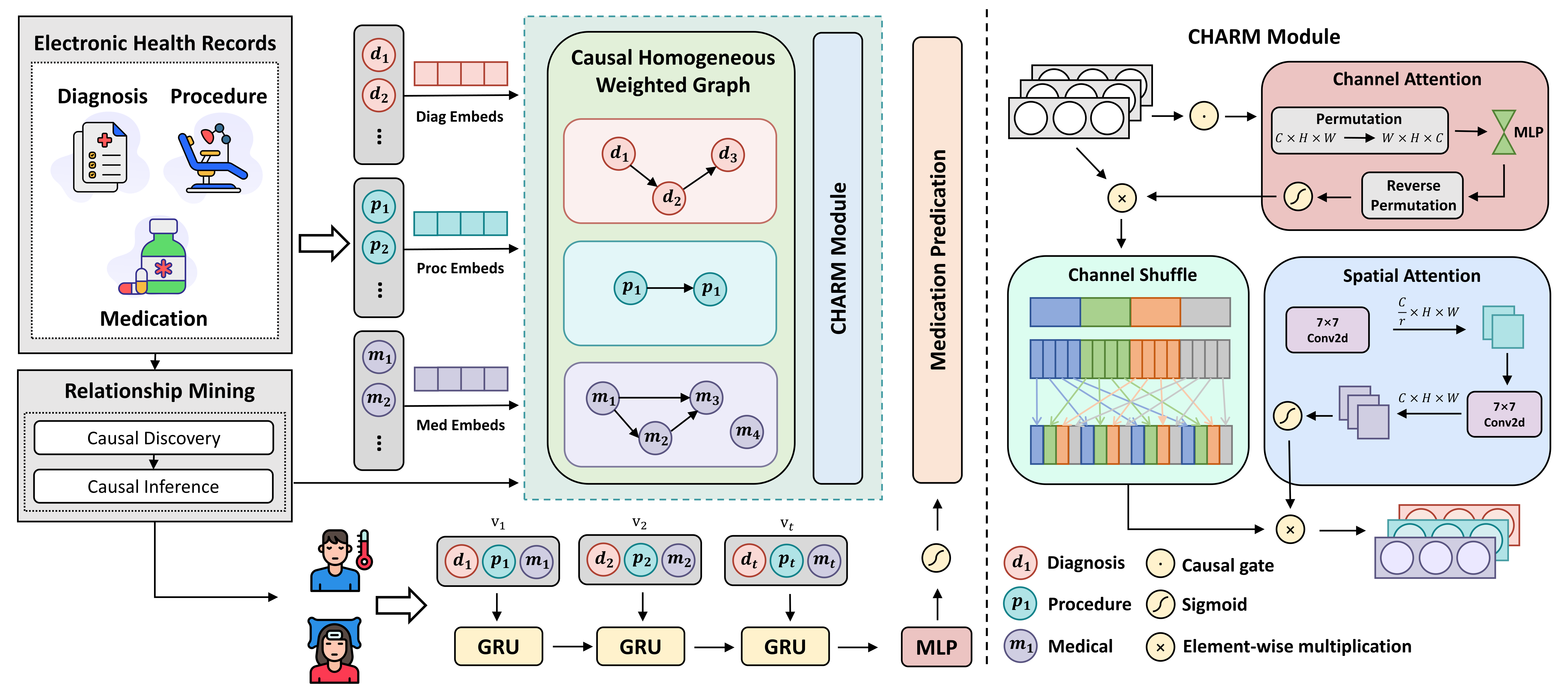}
  \caption{The overall architecture of the proposed CafeMed, comprising two key innovative components: (i) CWG that dynamically transforms causal effects into adaptive modulation weights for entity representations. (ii) The CHARM module that enables efficient cross-modal feature fusion through channel attention, channel shuffle, and spatial attention mechanisms.}
  \label{fig:teaser}
\end{figure*}

Our main contributions can be summarized as follows:
\begin{itemize}
    \item We pioneer the integration of causal inference with attention mechanisms for patient-centric medication recommendation, enabling dynamic adaptation of feature representations across diverse patient profiles.
    \item We introduce the CWG mechanism, which adjusts feature embeddings based on the causal effect strengths of diagnostic and procedural information on medication outcomes, effectively filtering drug conflicts and reducing adverse reaction risks.
    \item We propose the CHARM module to enhance synergistic representation among diagnoses, procedures, and medication history through multi-channel and spatial attention, improving predictive accuracy in a comprehensive manner.
\end{itemize}

\section{Related Works}
\subsection{Medication Recommendation}
Early medication recommendation approaches primarily relied on statistical association analysis or heuristic rules, identifying drug co-occurrence patterns from historical data. However, these methods often conflate correlation with causation, introducing spurious associations that compromise recommendation safety \cite{wu2022conditional}.
The advent of deep learning has significantly advanced this field. RETAIN \cite{choi2016retain} employs reverse-time attention mechanisms to identify critical clinical events in patient histories. LEAP \cite{zhang2017leap} formulates medication recommendation as sequential decision-making using reinforcement learning. GAMENet \cite{shang2019gamenet} leverages graph neural networks to capture drug interactions, while SafeDrug \cite{yang2021safedrug} incorporates molecular information to mitigate adverse drug interactions. Recent work like MoleRec \cite{yang2023molerec} utilizes molecular substructure information, and CausalMed \cite{li2024causalmed} introduces causal inference to discover genuine relationships among medical entities through directed acyclic graphs. Despite these advances, most existing models treat diseases and procedures as independent features, failing to capture their complex interdependencies and dynamic interactions in patient-specific contexts.
\subsection{Causal Inference in Healthcare}
Causal inference has emerged as a transformative paradigm in recommender systems, addressing fundamental challenges of confounding bias and spurious correlations inherent in observational data \cite{gao2024causal, yao2021survey}. Unlike traditional correlation-based approaches, causal methods explicitly model the underlying mechanisms driving user behaviors and outcomes, thereby enhancing both recommendation accuracy and interpretability.
The application of causal inference in recommender systems encompasses several key methodologies. Counterfactual reasoning approaches, such as those proposed by Zhang et al. \cite{zhang2021counterfactual}, employ inverse probability weighting and counterfactual prediction to mitigate selection bias and popularity bias. Causal representation learning methods \cite{wang2022causal} focus on disentangling causal factors from confounding variables to achieve more robust out-of-distribution recommendations. Additionally, causal discovery techniques \cite{si2022model} leverage constraint-based and score-based algorithms to uncover latent causal structures from observational data.
Healthcare applications present unique challenges for causal inference due to complex confounding relationships and treatment effect heterogeneity. Recent medication recommender systems have increasingly adopted causal frameworks to address these challenges and improve clinical decision accuracy. However, existing causal approaches in medication recommendation exhibit critical limitations: (1) static causal modeling that treats all relationships uniformly, and (2) lack of adaptive mechanisms to adjust causal relationship importance based on patient-specific states.
CafeMed addresses these fundamental challenges by introducing dynamic causal weight gating that adaptively modulates relationship strengths according to patient contexts, coupled with attention-based feature fusion that captures complex disease--procedure synergies. 

\section{Problem Formulation}
Given a patient’s longitudinal visit sequence $P = \{v_1, v_2, \ldots, v_t\}$, where
$v_t = \langle D_t, P_t, M_t \rangle$ denotes the diagnosis set, procedure set, and medication set at the $t$-th visit. Our goal is to learn a mapping $f:\{D_1^t,\; P_1^t,\; M_1^{t-1}\} \rightarrow M_t,$ which recommends an accurate and safe medication set $M_t$ under drug-–drug interaction constraints.

\section{Methods}
Building upon the causal inference framework, CafeMed introduces two key innovations: (i) Dynamic Causal Modulation via CWG that transforms static causal effects into adaptive feature weights, and (ii) Cross-modal Feature Fusion via CHARM that efficiently integrates multi-source medical information. Fig. 2 shows an overview of the proposed architecture.

\subsection{Causal Weight Generator (CWG)}
CWG transforms static causal effects into dynamic modulation weights for embedding enhancement. We employ the GIES algorithm \cite{nandy2018understanding, kalainathan2020causal} to construct causal graphs and quantify causal effects between medical entities. For each entity, we aggregate its causal impacts on medications:
\begin{equation}
\bar{\tau}_i = \frac{1}{|M|}\sum_{j=1}^{|M|} \tau_{ij},
\end{equation}
where $\tau_{ij}$ represents the causal effect and $|M|$ is the number of medications.

A lightweight neural network maps this scalar to a $d$-dimensional modulation vector:
\begin{equation}
\mathbf{w}_i = \text{Linear}_{d/4 \to d}(\text{ReLU}(\text{Linear}_{1 \to d/4}([\bar{\tau}_i]))),
\end{equation}
where the two linear layers progressively expand the dimension from 1 to $d$.

The entity embeddings are then modulated using a gating mechanism:
\begin{equation}
\mathbf{h}_i' = \mathbf{h}_i \odot (1 + \alpha \cdot \sigma(\mathbf{w}_i)),
\end{equation}
where $\alpha = 0.5$ controls the modulation strength, ensuring that causal information enhances rather than overwrites the original embeddings.

\subsection{CHARM Module}
CHARM enables effective cross-modal fusion through a novel combination of channel attention, channel shuffling, and spatial attention mechanisms. It processes concatenated multi-modal features to capture complex inter-modal relationships. The module also incorporates causal information through a gating mechanism, ensuring that medical causal relationships guide the attention process.

\subsubsection{Channel Attention}
The channel attention mechanism identifies and emphasizes the most informative feature channels across all modalities. Given input tensor $\mathbf{X} \in \mathbb{R}^{B \times C \times H \times W}$, we first apply permutation to rearrange dimensions:
\begin{equation}
\mathbf{X}_{\text{perm}} = \text{Permute}(\mathbf{X}, [B, H, W, C]) \to [B, H \times W, C].
\end{equation}

The permuted features are processed through a multi-layer perceptron (MLP) with bottleneck architecture:
\begin{equation}
\mathbf{F}_{\text{mlp}} = \text{Linear}_{C \to C}(\text{ReLU}(\text{Linear}_{C \to C/r}(\mathbf{X}_{\text{perm}}))),
\end{equation}
where $r=4$ is the reduction ratio.

After reverse permutation and sigmoid activation, we obtain the channel attention map:
\begin{equation}
\mathbf{A}_{\text{channel}} = \sigma(\text{Permute}^{-1}(\mathbf{F}_{\text{mlp}})).
\end{equation}

The attention is applied through element-wise multiplication:
\begin{equation}
\mathbf{X}_{\text{channel}} = \mathbf{X} \odot \mathbf{A}_{\text{channel}}.
\end{equation}

\subsubsection{Channel Shuffle}
To enhance information exchange between different modality groups, we introduce a channel shuffling operation. When the number of channels is divisible by 4, we partition channels into groups and shuffle them:
\begin{equation}
\mathbf{X}_{\text{shuffle}} = \text{ChannelShuffle}(\mathbf{X}_{\text{channel}}, g=4),
\end{equation}
where the shuffling operation rearranges channels as:
\begin{align}
[B, C, H, W] &\to [B, g, C/g, H, W] \to \notag \\
&\to [B, C/g, g, H, W] \to [B, C, H, W].
\end{align}

This shuffling breaks the rigid boundaries between modality-specific channels, promoting cross-modal feature interaction.

\subsubsection{Spatial Attention}
The spatial attention mechanism identifies important spatial regions across the feature map. We employ a dual-convolution architecture with instance normalization:
\begin{equation}
\mathbf{F}_{\text{spatial}} = \text{Conv}_{7\times7}(\text{IN}(\text{ReLU}(\text{Conv}_{7\times7}(\mathbf{X}_{\text{shuffle}})))),
\end{equation}
\begin{equation}
\mathbf{A}_{\text{spatial}} = \sigma(\text{IN}(\mathbf{F}_{\text{spatial}})),
\end{equation}
where $\text{Conv}_{7\times7}$ denotes convolution with kernel size 7 and padding 3, and IN is instance normalization.

The final output combines both attention mechanisms:
\begin{equation}
\mathbf{X}_{\text{output}} = \mathbf{X}_{\text{shuffle}} \odot \mathbf{A}_{\text{spatial}}.
\end{equation}

\subsubsection{Integration with CWG}
CHARM processes the causally-modulated embeddings from all three modalities simultaneously. The input features $\mathbf{X}_{\text{combined}} = [\mathbf{h}'_{\text{diag}} \Vert \mathbf{h}'_{\text{proc}} \Vert \mathbf{h}'_{\text{med}}]$ are first aligned and reshaped for 2D processing. Within CHARM, a causal gate mechanism further modulates the features:
\begin{equation}
\mathbf{X}_{\text{gated}} = \mathbf{X} \odot \sigma(\text{CausalGate}),
\end{equation}
where the causal gate is derived from the aggregated causal effects. This dual-level causal integration ensures that both embedding-level (through CWG) and attention-level (through causal gate) causal information guides the feature fusion process.

\subsection{Sequence Modeling and Medication Prediction}
We employ three separate GRU encoders to capture temporal dependencies:
\begin{align}
\mathbf{o}^{(m)}_t, \mathbf{h}^{(m)}_t &= \text{GRU}^{(m)}(\mathbf{x}^{(m)}_t, \mathbf{h}^{(m)}_{t-1}), \notag \\
& \quad m \in \{\text{diag}, \text{proc}, \text{med}\}.
\end{align}

Additionally, we utilize homogeneous graph networks to capture within-modality relationships. The final representation for each modality combines both homogeneous and heterogeneous (CHARM-processed) features:
\begin{equation}
\mathbf{e}^{(m)} = \rho^{(m)}_0 \cdot \mathbf{e}^{(m)}_{\text{homo}} + \rho^{(m)}_1 \cdot \mathbf{e}^{(m)}_{\text{CHARM}},
\end{equation}
where $\rho^{(m)}$ are learnable fusion weights.

The patient representation concatenates both hidden states and final outputs:
\begin{equation}
\mathbf{z}_{\text{patient}} = [\mathbf{h}^{\text{diag}}_{T} \Vert \mathbf{h}^{\text{proc}}_{T} \Vert \mathbf{h}^{\text{med}}_{T} \Vert \mathbf{o}^{\text{diag}}_{T} \Vert \mathbf{o}^{\text{proc}}_{T} \Vert \mathbf{o}^{\text{med}}_{T}],
\end{equation}
which is then mapped to medication scores through a query network. The model is trained with a combined objective that balances prediction accuracy (binary cross-entropy) with drug--drug interaction (DDI) safety constraints ($\beta = 0.0005$), ensuring both effective and safe medication recommendations.

\begin{table}[t]
\renewcommand{\arraystretch}{1.2}
\caption{Statistics of the datasets.}
\label{tab:dataset_stats}
\centering
\begin{tabular}{lcc}
\toprule
\textbf{Item} & \textbf{MIMIC-III} & \textbf{MIMIC-IV} \\
\midrule
\# patients & 6,350 & 60,125 \\
\# visits events & 15,032 & 156,810 \\
\# diseases & 1,958 & 2,000 \\
\# procedures & 1,430 & 1,500 \\
\# medications & 131 & 131 \\
avg. \# of visits & 2.37 & 2.61 \\
avg. \# of medications & 11.44 & 6.66 \\
\bottomrule
\end{tabular}
\end{table}

\begin{table*}[!t]
\renewcommand{\arraystretch}{1.2}
\caption{Performance comparison of all models on the test set in terms of accuracy and safety metrics. The best and second-best results are highlighted in \textbf{bold} and \underline{underlined}, respectively, based on paired t-tests ($p < 0.05$).}
\label{tab:performance}
\centering
\begin{tabular}{lcccccccccc}
\toprule
\multirow{2}{*}{Model} & \multicolumn{5}{c}{\textbf{MIMIC-III}} & \multicolumn{5}{c}{\textbf{MIMIC-IV}} \\
\cmidrule(lr){2-6} \cmidrule(lr){7-11}
 & Jaccard↑ & DDI↓ & F1↑ & PRAUC↑ & Avg.\#Med & Jaccard↑ & DDI↓ & F1↑ & PRAUC↑ & Avg.\#Med \\
\midrule
LR & 0.4924 & 0.0830 & 0.6490 & 0.7548 & 16.0489 & 0.4569 & 0.0783 & 0.6064 & 0.6613 & 8.5746 \\
ECC & 0.4856 & 0.0817 & 0.6438 & 0.7590 & 16.2578 & 0.4327 & 0.0764 & 0.6129 & 0.6530 & 8.7934 \\
RETAIN & 0.4871 & 0.0879 & 0.6473 & 0.7600 & 19.4222 & 0.4234 & 0.0936 & 0.5785 & 0.6801 & 10.9576 \\
LEAP & 0.4526 & 0.0762 & 0.6147 & 0.6555 & 18.6240 & 0.4254 & 0.0688 & 0.5794 & 0.6659 & 11.3606 \\
GAMENet & 0.4994 & 0.0890 & 0.6560 & 0.7656 & 27.7703 & 0.4565 & 0.0898 & 0.6103 & 0.6829 & 18.5895 \\
SafeDrug & 0.5154 & \textbf{0.0655} & 0.6722 & 0.7627 & 19.4111 & 0.4487 & \textbf{0.0604} & 0.6014 & 0.6948 & 13.6943 \\
MICRON & 0.5219 & 0.0727 & 0.6761 & 0.7489 & 19.3505 & 0.4640 & 0.0691 & 0.6167 & 0.6919 & 12.7701 \\
COGNet & 0.5312 & 0.0839 & 0.6744 & 0.7708 & 27.6335 & 0.4775 & 0.0911 & 0.6233 & 0.6524 & 18.7235 \\
MoleRec & 0.5293 & 0.0726 & 0.6834 & 0.7746 & 22.0125 & 0.4744 & 0.0722 & 0.6262 & 0.7124 & 13.4806 \\
CausalMed & \underline{0.5389} & 0.0709 & \underline{0.6916} & \underline{0.7826} & 20.5419 & \underline{0.4899} & \underline{0.0677} & \underline{0.6412} & \underline{0.7338} & 14.4295 \\
\midrule
CafeMed & \textbf{0.5431} & \underline{0.0708} & \textbf{0.6958} & \textbf{0.7871} & 20.9509 & \textbf{0.4946} & 0.0697 & \textbf{0.6449} & \textbf{0.7386} & 14.3829 \\
\bottomrule
\end{tabular}
\end{table*}

\begin{table*}[!t]
\renewcommand{\arraystretch}{1.2}
\caption{Ablation results of CafeMed variants. Best and second-best performances are marked in \textbf{bold} and \underline{underlined}, respectively.}
\label{tab:causalmed_variants}
\centering
\begin{tabular}{lccccccccc}
\toprule
\multirow{2}{*}{Model} & \multicolumn{4}{c}{MIMIC-III} & \multicolumn{4}{c}{MIMIC-IV} \\
\cmidrule(lr){2-5} \cmidrule(lr){6-9}
 & Jaccard↑ & DDI↓ & F1↑ & PRAUC↑ & Jaccard↑ & DDI↓ & F1↑ & PRAUC↑ \\
\midrule
w/o CWG & \underline{0.5402} & 0.0740 & \underline{0.6932} & \underline{0.7870} & \underline{0.4931} & 0.0702 & \underline{0.6437} & \underline{0.7383} \\
w/o CHARM & 0.5342 & \underline{0.0713} & 0.6878 & 0.7815 & 0.4859 & \textbf{0.0674} & 0.6360 & 0.7293 \\
w/o Full & 0.5332 & 0.0720 & 0.6869 & 0.7817 & 0.4849 & 0.0703 & 0.6360 & 0.7294 \\
\midrule
\textbf{CafeMed} & \textbf{0.5431} & \textbf{0.0708} & \textbf{0.6958} & \textbf{0.7871} & \textbf{0.4946} & \underline{0.0697} & \textbf{0.6449} & \textbf{0.7386} \\
\bottomrule
\end{tabular}
\end{table*}

\section{Experiments}
We conduct a series of experiments to address the following research
questions:
\begin{itemize}
\item {\texttt{\textbf{RQ1:}}} How effective is the proposed CafeMed architecture compared with state-of-the-art medication recommendation models?
\item {\texttt{\textbf{RQ2:}}} How do the key components within CafeMed contribute to its overall performance?
\item {\texttt{\textbf{RQ3:}}} How does CafeMed improve computational efficiency while enhancing recommendation quality?
\end{itemize}

\subsection{Experimental Settings}
\subsubsection{Datasets}
We evaluate CafeMed on MIMIC-III \cite{johnson2016mimic} and MIMIC-IV \cite{johnson2023mimic}, following standard preprocessing procedures \cite{yang2021safedrug}. The datasets are split into training, validation, and test sets with a 4:1:1 ratio. We perform bootstrap sampling with ten rounds on test sets to ensure statistical significance. The detailed statistics of the processed datasets are presented in the Table I.

\subsubsection{Evaluation Metrics}
We employ five metrics for comprehensive evaluation: (1) \textbf{Jaccard} measures the overlap between predicted and actual medication sets; (2) \textbf{DDI rate} quantifies the proportion of harmful drug interactions; (3) \textbf{F1-score} balances precision and recall; (4) \textbf{PRAUC} evaluates performance under class imbalance; (5) \textbf{Avg.\#Med} indicates the average number of recommended medications.

\subsubsection{Baselines}
We compare with ten representative methods: traditional approaches (LR \cite{indra2016using}, ECC \cite{read2011classifier}), deep learning models (RETAIN \cite{choi2016retain}, LEAP \cite{zhang2017leap}, GAMENet \cite{shang2019gamenet}), molecule-aware methods (SafeDrug \cite{yang2021safedrug}, MICRON \cite{yang2021change}, COGNet \cite{wu2022conditional}, MoleRec \cite{yang2023molerec}), and causal inference approach (CausalMed \cite{li2024causalmed}).

\subsubsection{Implementation Details}
Following established benchmarks, we set embedding dimension to 64, learning rate to $5 \times 10^{-4}$ with Adam optimizer, and dropout rate to 0.7. The DDI threshold is constrained to 0.06 with $L_2$ regularization coefficient of 0.005. All experiments are conducted on a platform with 90GB RAM, 12-core CPU, and NVIDIA RTX 3080Ti GPU.

\subsection{Performance Comparison (RQ1)}
As shown in Table II, we evaluate CafeMed against state-of-the-art baselines to demonstrate its effectiveness. Traditional methods LR and ECC fail to model temporal dependencies and drug interactions, leading to inferior performance across all metrics. Deep learning models RETAIN and LEAP enhance accuracy through sequential modeling but neglect safety constraints, resulting in elevated DDI risks. GAMENet achieves high accuracy but produces the highest DDI rate due to excessive medication recommendations without safety considerations. SafeDrug prioritizes DDI reduction through molecular information but substantially compromises accuracy. MICRON employs residual networks for patient representation yet maintains high DDI rates. COGNet and MoleRec achieve better balance but remain suboptimal in handling drug interaction complexities. CausalMed introduces causal inference to capture medical entity relationships, improving accuracy significantly. However, its static causal modeling lacks adaptive mechanisms for patient-specific contexts.

In contrast, CafeMed addresses these limitations through dynamic causal weight modulation via CWG and cross-modal attention fusion via CHARM. CWG adaptively adjusts causal relationship strengths based on individual patient states, while CHARM captures synergistic effects between diagnoses and procedures. This design enables CafeMed to achieve superior accuracy while maintaining the lower DDI rates across both datasets. The consistent improvements validate that our approach effectively balances recommendation accuracy with medication safety, establishing a new benchmark for safe and personalized medication recommendation.

\subsection{Ablation Studies (RQ2)}
As illustrated in Table III, to investigate the contribution of each component in CafeMed, we conduct ablation experiments with three variants: w/o CWG, w/o CHARM, and w/o Full (both components removed).

The absence of CWG eliminates dynamic causal modulation, forcing the model to rely on static relationships. This results in degraded performance across all metrics, particularly in DDI rates, demonstrating that adaptive causal weight adjustment based on patient states is essential for both accuracy and safety. The exclusion of CHARM, which is replaced by simple concatenation, leads to the most substantial performance drop. This confirms that capturing cross-modal synergies between diagnoses and procedures through sophisticated attention mechanisms is crucial for understanding their joint influence on medication decisions.

The baseline variant without both components exhibits the worst performance, highlighting the complementary nature of CWG and CHARM. The ablation results reveal that: (1) Both components contribute significantly, with CHARM showing slightly larger impact on accuracy; (2) The full model achieves optimal balance between recommendation accuracy and medication safety through their synergistic integration; (3) Consistent patterns across both datasets validate the robustness of our design. These findings confirm that dynamic causal reasoning and cross-modal attention jointly enable CafeMed to achieve superior performance.

\begin{table}[b]
\renewcommand{\arraystretch}{1.2}
\caption{Comparison of training and inference times for CausalMed and CafeMed.}
\label{tab:model_comparison}
\centering
\begin{tabular}{lcccc}
\toprule
\textbf{Method} & \textbf{Conv.} & \textbf{Time/epoch} & \textbf{Total time} & \textbf{Inference} \\
                & \textbf{epoch} & \textbf{(s)}        & \textbf{(s)}        & \textbf{(s)} \\
\midrule
CausalMed & 36 & 1,007.93 & 36,057.93 & 108.94 \\
CafeMed & \textbf{16} & \textbf{633.49} & \textbf{10,135.84} & \textbf{65.05} \\
\midrule
\textit{Improv. (\%)} & \textit{55.56} & \textit{37.15} & \textit{71.89} & \textit{40.29} \\
\bottomrule
\end{tabular}
\end{table}

% \begin{table}[b]
% \renewcommand{\arraystretch}{1.2}
% \caption{Comparison of training and inference times for CausalMed and CafeMed.}
% \label{tab:model_comparison}
% \centering
% \begin{tabular}{lcccc}
% \toprule
% \textbf{Method} & \textbf{Conv.} & \textbf{Time/epoch} & \textbf{Total time} & \textbf{Inference} \\
%                 & \textbf{epoch} & \textbf{(s)}        & \textbf{(s)}        & \textbf{(s)} \\
% \midrule
% CausalMed & \cellcolor{gray!15}36 & \cellcolor{gray!15}1,007.93 & \cellcolor{gray!15}36,057.93 & \cellcolor{gray!15}108.94 \\
% CafeMed & \cellcolor{green!15}\textbf{16} & \cellcolor{green!15}\textbf{633.49} & \cellcolor{green!15}\textbf{10,135.84} & \cellcolor{green!15}\textbf{65.05} \\
% \midrule
% \textit{Improv. (\%)} & \cellcolor{red!15}\textit{55.56} & \cellcolor{red!15}\textit{37.15} & \cellcolor{red!15}\textit{71.89} & \cellcolor{red!15}\textit{40.29} \\
% \bottomrule
% \end{tabular}
% \end{table}

\subsection{Efficiency Analysis (RQ3)}
Table IV compares computational efficiency between CafeMed and CausalMed. To eliminate randomness effects, we conducted performance tests on two identical but independent servers and report averaged results. Despite incorporating additional modules (CWG and CHARM), CafeMed achieves faster convergence and reduced inference time. This efficiency gain stems from our lightweight design: CWG employs a simple two-layer network for causal weight transformation instead of complex graph computations, while CHARM efficiently processes concatenated features through unified attention rather than separate modal processing. The streamlined architecture eliminates redundant causal graph traversals during inference, enabling practical deployment in resource-constrained settings while maintaining superior recommendation quality.

\section{Conclusion}
We present CafeMed, a novel framework that integrates dynamic causal reasoning with cross-modal attention for safe and accurate medication recommendation. By introducing the Causal Weight Generator (CWG) for adaptive relationship modulation and the CHARM module for synergistic feature fusion, CafeMed effectively captures the complex interdependencies between diagnoses, procedures, and medications. Comprehensive experiments on real-world clinical datasets demonstrate the efficacy and robustness of our proposed framework.

\section*{Acknowledgments}
This work was partly supported by Institute of Information \& communications Technology Planning \& Evaluation (IITP) grant funded by the Korea government (MSIT) (No.RS-2022-00155885, Artificial Intelligence Convergence Innovation Human Resources Development (Hanyang University ERICA)) and the MSIT (Ministry of Science and ICT), Korea, under the Convergence security core talent training business support program (IITP-2024-RS-2024-00423071) supervised by the IITP(Institute of Information \& Communications Technology Planning \& Evaluation) and the MSIT (Ministry of Science, ICT), Korea, under the National Program for Excellence in SW),  supervised by the IITP (Institute of Information \& communications Technology Planing \& Evaluation) in 2025(2024-0-00058).

% \section*{References}

\bibliography{references}

@article{yang2021safedrug,
  title={Safedrug: Dual molecular graph encoders for recommending effective and safe drug combinations},
  author={Yang, Chaoqi and Xiao, Cao and Ma, Fenglong and Glass, Lucas and Sun, Jimeng},
  journal={arXiv preprint arXiv:2105.02711},
  year={2021}
}

@inproceedings{chen2023context,
  title={Context-aware safe medication recommendations with molecular graph and DDI graph embedding},
  author={Chen, Qianyu and Li, Xin and Geng, Kunnan and Wang, Mingzhong},
  booktitle={Proceedings of the AAAI conference on artificial intelligence},
  volume={37},
  number={6},
  pages={7053--7060},
  year={2023}
}

@article{li2024stratmed,
  title={StratMed: Relevance stratification between biomedical entities for sparsity on medication recommendation},
  author={Li, Xiang and Liang, Shunpan and Hou, Yulei and Ma, Tengfei},
  journal={Knowledge-Based Systems},
  volume={284},
  pages={111239},
  year={2024},
  publisher={Elsevier}
}

@article{cowie2017electronic,
  title={Electronic health records to facilitate clinical research},
  author={Cowie, Martin R and Blomster, Juuso I and Curtis, Lesley H and Duclaux, Sylvie and Ford, Ian and Fritz, Fleur and Goldman, Samantha and Janmohamed, Salim and Kreuzer, J{\"o}rg and Leenay, Mark and others},
  journal={Clinical Research in Cardiology},
  volume={106},
  number={1},
  pages={1--9},
  year={2017},
  publisher={Springer}
}

@article{evans2016electronic,
  title={Electronic health records: then, now, and in the future},
  author={Evans, R Scott},
  journal={Yearbook of medical informatics},
  volume={25},
  number={S 01},
  pages={S48--S61},
  year={2016},
  publisher={Georg Thieme Verlag KG}
}

@inproceedings{ma2018general,
  title={A general framework for diagnosis prediction via incorporating medical code descriptions},
  author={Ma, Fenglong and Wang, Yaqing and Xiao, Houping and Yuan, Ye and Chitta, Radha and Zhou, Jing and Gao, Jing},
  booktitle={2018 IEEE International Conference on Bioinformatics and Biomedicine (BIBM)},
  pages={1070--1075},
  year={2018},
  organization={IEEE}
}

@inproceedings{symeonidis2021recommending,
  title={Recommending what drug to prescribe next for accurate and explainable medical decisions},
  author={Symeonidis, Panagiotis and Chairistanidis, Stergios and Zanker, Markus},
  booktitle={2021 IEEE 34th International Symposium on Computer-Based Medical Systems (CBMS)},
  pages={213--218},
  year={2021},
  organization={IEEE}
}

@inproceedings{wang2021learning,
  title={Learning intents behind interactions with knowledge graph for recommendation},
  author={Wang, Xiang and Huang, Tinglin and Wang, Dingxian and Yuan, Yancheng and Liu, Zhenguang and He, Xiangnan and Chua, Tat-Seng},
  booktitle={Proceedings of the web conference 2021},
  pages={878--887},
  year={2021}
}

@inproceedings{shang2018knowledge,
  title={Knowledge guided multi-instance multi-label learning via neural networks in medicines prediction},
  author={Shang, Junyuan and Hong, Shenda and Zhou, Yuxi and Wu, Meng and Li, Hongyan},
  booktitle={Asian Conference on Machine Learning},
  pages={831--846},
  year={2018},
  organization={PMLR}
}

@inproceedings{wang2019order,
  title={Order-free medicine combination prediction with graph convolutional reinforcement learning},
  author={Wang, Shanshan and Ren, Pengjie and Chen, Zhumin and Ren, Zhaochun and Ma, Jun and de Rijke, Maarten},
  booktitle={Proceedings of the 28th ACM international conference on information and knowledge management},
  pages={1623--1632},
  year={2019}
}

@inproceedings{wang2022ffbdnet,
  title={FFBDNet: Feature fusion and bipartite decision networks for recommending medication combination},
  author={Wang, Zisen and Liang, Ying and Liu, Zhengjun},
  booktitle={Joint European conference on machine learning and knowledge discovery in databases},
  pages={419--436},
  year={2022},
  organization={Springer}
}

@inproceedings{li2024causalmed,
  title={Causalmed: Causality-based personalized medication recommendation centered on patient health state},
  author={Li, Xiang and Liang, Shunpan and Lei, Yu and Li, Chen and Hou, Yulei and Zheng, Dashun and Ma, Tengfei},
  booktitle={Proceedings of the 33rd ACM international conference on information and knowledge management},
  pages={1276--1285},
  year={2024}
}

@article{lee2020clinical,
  title={Clinical applications of continual learning machine learning},
  author={Lee, Cecilia S and Lee, Aaron Y},
  journal={The Lancet Digital Health},
  volume={2},
  number={6},
  pages={e279--e281},
  year={2020},
  publisher={Elsevier}
}

@inproceedings{shang2019gamenet,
  title={Gamenet: Graph augmented memory networks for recommending medication combination},
  author={Shang, Junyuan and Xiao, Cao and Ma, Tengfei and Li, Hongyan and Sun, Jimeng},
  booktitle={proceedings of the AAAI Conference on Artificial Intelligence},
  volume={33},
  number={01},
  pages={1126--1133},
  year={2019}
}

@inproceedings{wu2022conditional,
  title={Conditional generation net for medication recommendation},
  author={Wu, Rui and Qiu, Zhaopeng and Jiang, Jiacheng and Qi, Guilin and Wu, Xian},
  booktitle={Proceedings of the ACM web conference 2022},
  pages={935--945},
  year={2022}
}

@inproceedings{zhang2017leap,
  title={LEAP: learning to prescribe effective and safe treatment combinations for multimorbidity},
  author={Zhang, Yutao and Chen, Robert and Tang, Jie and Stewart, Walter F and Sun, Jimeng},
  booktitle={proceedings of the 23rd ACM SIGKDD international conference on knowledge Discovery and data Mining},
  pages={1315--1324},
  year={2017}
}

@article{choi2016retain,
  title={Retain: An interpretable predictive model for healthcare using reverse time attention mechanism},
  author={Choi, Edward and Bahadori, Mohammad Taha and Sun, Jimeng and Kulas, Joshua and Schuetz, Andy and Stewart, Walter},
  journal={Advances in neural information processing systems},
  volume={29},
  year={2016}
}

@inproceedings{yang2023molerec,
  title={Molerec: Combinatorial drug recommendation with substructure-aware molecular representation learning},
  author={Yang, Nianzu and Zeng, Kaipeng and Wu, Qitian and Yan, Junchi},
  booktitle={Proceedings of the ACM web conference 2023},
  pages={4075--4085},
  year={2023}
}

@article{gao2024causal,
  title={Causal inference in recommender systems: A survey and future directions},
  author={Gao, Chen and Zheng, Yu and Wang, Wenjie and Feng, Fuli and He, Xiangnan and Li, Yong},
  journal={ACM Transactions on Information Systems},
  volume={42},
  number={4},
  pages={1--32},
  year={2024},
  publisher={ACM New York, NY}
}

@article{yao2021survey,
  title={A survey on causal inference},
  author={Yao, Liuyi and Chu, Zhixuan and Li, Sheng and Li, Yaliang and Gao, Jing and Zhang, Aidong},
  journal={ACM Transactions on Knowledge Discovery from Data (TKDD)},
  volume={15},
  number={5},
  pages={1--46},
  year={2021},
  publisher={ACM New York, NY, USA}
}

@inproceedings{zhang2021counterfactual,
  title={Counterfactual reward modification for streaming recommendation with delayed feedback},
  author={Zhang, Xiao and Jia, Haonan and Su, Hanjing and Wang, Wenhan and Xu, Jun and Wen, Ji-Rong},
  booktitle={Proceedings of the 44th international ACM SIGIR conference on research and development in information retrieval},
  pages={41--50},
  year={2021}
}

@inproceedings{wang2022causal,
  title={Causal representation learning for out-of-distribution recommendation},
  author={Wang, Wenjie and Lin, Xinyu and Feng, Fuli and He, Xiangnan and Lin, Min and Chua, Tat-Seng},
  booktitle={Proceedings of the ACM Web Conference 2022},
  pages={3562--3571},
  year={2022}
}

@inproceedings{si2022model,
  title={A model-agnostic causal learning framework for recommendation using search data},
  author={Si, Zihua and Han, Xueran and Zhang, Xiao and Xu, Jun and Yin, Yue and Song, Yang and Wen, Ji-Rong},
  booktitle={Proceedings of the ACM web conference 2022},
  pages={224--233},
  year={2022}
}

@article{nandy2018understanding,
  title={Understanding consistency in hybrid causal structure learning},
  author={Nandy, Preetam and Hauser, Alain and Maathuis, Marloes H},
  journal={Ann. Stat},
  year={2018}
}

@article{kalainathan2020causal,
  title={Causal discovery toolbox: Uncovering causal relationships in python},
  author={Kalainathan, Diviyan and Goudet, Olivier and Dutta, Ritik},
  journal={Journal of Machine Learning Research},
  volume={21},
  number={37},
  pages={1--5},
  year={2020}
}

@article{johnson2016mimic,
  title={MIMIC-III, a freely accessible critical care database},
  author={Johnson, Alistair EW and Pollard, Tom J and Shen, Lu and Lehman, Li-wei H and Feng, Mengling and Ghassemi, Mohammad and Moody, Benjamin and Szolovits, Peter and Anthony Celi, Leo and Mark, Roger G},
  journal={Scientific data},
  volume={3},
  number={1},
  pages={1--9},
  year={2016},
  publisher={Nature Publishing Group}
}

@article{johnson2023mimic,
  title={MIMIC-IV, a freely accessible electronic health record dataset},
  author={Johnson, Alistair EW and Bulgarelli, Lucas and Shen, Lu and Gayles, Alvin and Shammout, Ayad and Horng, Steven and Pollard, Tom J and Hao, Sicheng and Moody, Benjamin and Gow, Brian and others},
  journal={Scientific data},
  volume={10},
  number={1},
  pages={1},
  year={2023},
  publisher={Nature Publishing Group UK London}
}

@inproceedings{indra2016using,
  title={Using logistic regression method to classify tweets into the selected topics},
  author={Indra, ST and Wikarsa, Liza and Turang, Rinaldo},
  booktitle={2016 international conference on advanced computer science and information systems (icacsis)},
  pages={385--390},
  year={2016},
  organization={IEEE}
}

@article{read2011classifier,
  title={Classifier chains for multi-label classification},
  author={Read, Jesse and Pfahringer, Bernhard and Holmes, Geoff and Frank, Eibe},
  journal={Machine learning},
  volume={85},
  number={3},
  pages={333--359},
  year={2011},
  publisher={Springer}
}

@article{yang2021change,
  title={Change matters: Medication change prediction with recurrent residual networks},
  author={Yang, Chaoqi and Xiao, Cao and Glass, Lucas and Sun, Jimeng},
  journal={arXiv preprint arXiv:2105.01876},
  year={2021}
}
\bibliographystyle{ieeetr}
% \bibitem{b1} Martin R Cowie, Juuso I Blomster, Lesley H Curtis, Sylvie Duclaux, Ian Ford, Fleur Fritz, Samantha Goldman, Salim Janmohamed, Jörg Kreuzer, Mark Leenay, et al. 2017. Electronic health records to facilitate clinical research. Clinical Research in Cardiology 106, 1 (2017), 1–9.
% \bibitem{b2} R Scott Evans. 2016. Electronic health records: then, now, and in the future. Yearbook of medical informatics 25, S 01 (2016), S48–S61.

\vspace{12pt}

\end{document}